\documentclass[letterpaper, 10 pt, conference]{ieeeconf}  

\let\labelindent\relax
\usepackage{enumitem}
\usepackage{graphicx}
\usepackage{subcaption}
\usepackage{caption}
\usepackage{xcolor}
\usepackage{amsmath}

\IEEEoverridecommandlockouts                              

\title{\LARGE \bf
Multi-Camera Worker Tracking in Logistics Warehouse \\Considering Wide-Angle Distortion
}

\author{
    Yuki Mori$^{*,\ddagger}$, 
    Kazuma Kano$^{*}$,
    Yusuke Asai$^{*}$,\\
    Shin Katayama$^{*}$,
    Kenta Urano$^{*}$, 
    Takuro Yonezawa$^{*}$, 
    Nobuo Kawaguchi$^{*,\dagger}$
\thanks{$^{*}$Graduate School of Engineering, Nagoya University, Nagoya, Japan}
\thanks{$^{\dagger}$Institutes of Innovation for Future Society, Nagoya University, Nagoya, Japan}
\thanks{$^{\ddagger}$email: ymori@ucl.nuee.nagoya-u.ac.jp}
}

\begin{document}

\maketitle
\thispagestyle{empty}
\pagestyle{empty}

\begin{abstract}
With the spread of e-commerce, the logistics market is growing around the world. 
Therefore, improving the efficiency of warehouse operations is essential.
To achieve this, various approaches have been explored, and among them, the use of digital twins is gaining attention.
To make this approach possible, it is necessary to accurately collect the positions of workers in a warehouse and reflect them in a virtual space.
However, a single camera has limitations in its field of view, therefore sensing with multiple cameras is necessary.
In this study, we explored a method to track workers using 19 wide-angle cameras installed on the ceiling, looking down at the floor of the logistics warehouse.
To understand the relationship between the camera coordinates and the actual positions in the warehouse, we performed alignment based on the floor surface.
However, due to the characteristics of wide-angle cameras, significant distortion occurs at the edges of the image, particularly in the vertical direction.
To address this, the detected worker positions from each camera were aligned based on foot positions, reducing the effects of image distortion, and enabling accurate position alignment across cameras.
As a result, we confirmed an improvement of over 20\% in tracking accuracy.
Furthermore, we compared multiple methods for utilizing appearance features and validated the effectiveness of the proposed approach.

\end{abstract}

\section{INTRODUCTION}
\begin{figure*}[t]
    \centering
    \includegraphics[width=\textwidth]{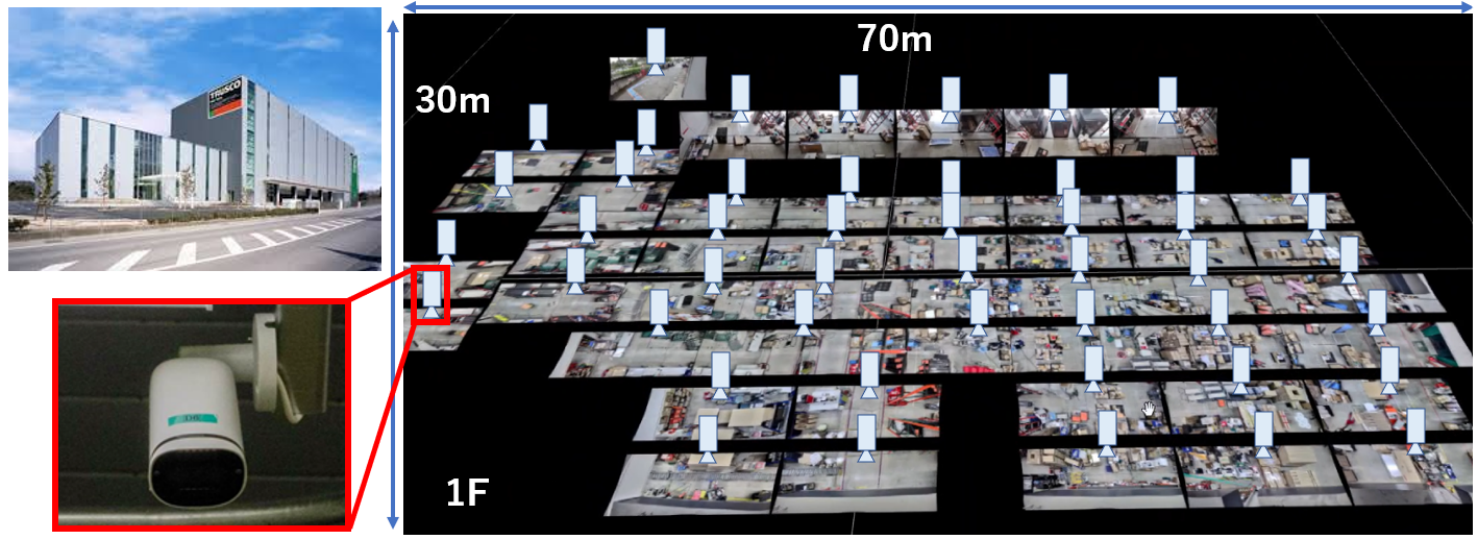}
    \caption{Large-scale camera infrastructure}
    \label{fig:cam_config_1st_floor}
\end{figure*}

With the spread of e-commerce, the logistics market is growing around the world. 
Therefore, workload inside warehouses is increasing, making the improvement of operational efficiency a critical challenge.
Various approaches are being explored to improve operational efficiency, including robot route optimization \cite{li2022optimal} and warehouse layout optimization \cite{hu2023commerce}.
Among them, the approach using digital twins has attracted much attention \cite{maddikunta2022industry}.
A digital twin is a virtual model of a real-world object, system, or process that allows simulations and other activities to be conducted in a virtual environment \cite{barricelli2019survey}.
By using digital twins, it is possible to run virtual experiments aimed at improving efficiency with low cost and minimal impact on the actual site.
However, in order to construct a digital twin, it is necessary to accurately digitize the physical space by sensing the environment.
In particular, accurately obtaining the positions of workers in logistics warehouses is essential.

As a method for acquiring location information, we have implemented a beacon-based approach by attaching beacons to workers \cite{kano2022smartphone}, but this method is difficult to apply to packages and tools.
In this regard, camera-based sensing does not require attaching additional devices to workers or packages, and can be relatively easy to apply to various targets.
However, a single camera has limitations in its field of view, therefore sensing with multiple cameras is necessary.

We built a large-scale camera system in a logistics warehouse, consisting of over 80 fixed cameras installed across five floors, including the first floor shown in Fig.\ref{fig:cam_config_1st_floor}.
In this study, we focus on 19 wide-angle cameras mounted on the ceiling of the first floor, capturing the floor from a top-down view, and propose a multi-camera tracking method for workers.
Specifically, we converted the positions of workers detected by each camera from its local coordinate system to a global coordinate system, which is a common warehouse-level coordinate system.
Then, by matching the tracking data from individual cameras, we enabled worker tracking throughout the entire receiving area of the warehouse.
In particular, we used the foot positions of workers instead of the centers of the detected bounding boxes (bbox), which have been widely adopted in existing approaches.
This idea helped reduce the effects of image distortion caused by wide-angle cameras—especially vertical distortion that is more noticeable near the edges of the image—and also reduced errors in camera alignment.
As a result, our method made it possible to track workers more accurately throughout the entire receiving area of the warehouse.
In addition, utilizing appearance features can further improve tracking accuracy.
However, in wide-angle camera footage, a person's appearance can vary significantly depending on their position in the frame, and objects in the warehouse environment often obscure parts of the body.
To address this, we compared two methods for utilizing appearance features—simple averaging and a method that considers both position and movement direction—and demonstrated their effectiveness in improving worker re-identification (ReID) accuracy across multiple cameras.

The contributions of this study are as follows:
\begin{itemize}
    \item We proposed a method that detects workers using 19 wide-angle cameras, and tracks them across the entire receiving area of the warehouse, achieving evaluation scores of HOTA 51.0, IDF1 54.7, and MOTA 79.7.
     \item We conducted a comparative evaluation between the use of center of bbox and foot position, and demonstrated that using foot position reduces the effects of distortion in wide-angle images and misalignment between multiple cameras, enabling more accurate tracking.
    \item We compared two appearance feature methods—simple averaging and one considering position and movement direction—clarifying their strengths and weaknesses and providing guidance for method selection based on the application scenario.
\end{itemize}

\section{RELATED WORK}
\subsection{Single-Camera Tracking}
In recent years, as object detection has become more accurate, the accuracy of multi-object tracking (MOT) using a single camera has also improved.
To support this progress, a variety of improved MOT methods have been proposed.
Among them, many approaches incorporate Kalman filters for object association \cite{zhang2022bytetrack, du2023strongsort, wang2024smiletrack}.
In particular, ByteTrack \cite{zhang2022bytetrack} achieves high-accuracy tracking by using low-confidence detection results as well as high-confidence ones, which were previously ignored in conventional methods.
Also, some methods have been proposed that train detection and tracking at the same time.
FairMOT \cite{zhang2021fairmot} is a one-shot tracker with two branches in one neural network: object detection and person re-identification (ReID). 
It achieved high performance by balancing both accuracies.
In addition, Transformer-based methods like TransTrack \cite{sun2020transtrack} use object features from the previous frame as queries to detect in the current frame. 
This framework matches new and past objects at the same time using attention.
Therefore, in single-camera MOT, many methods have been developed — from simple motion-based matching to approaches that use appearance features, end-to-end learning, and advanced association with Transformers.

\subsection{ReID:Re-identification}
ReID is a technique for identifying the same person from images taken by different cameras, such as in surveillance networks.
It is a core component of multi-camera tracking systems.
With the progress of deep learning, many models have been proposed to learn appearance features that represent individual identity.
Omni-Scale Network (OSNet) \cite{zhou2019omni} is a well-known example. 
It uses a lightweight CNN structure to capture features at different scales and achieves state-of-the-art accuracy on several ReID benchmarks.
In addition, He et al. \cite{he2021transreid} proposed a pioneering Transformer-based method, which outperformed many CNN-based approaches on various benchmarks.
Furthermore, Luo et al. \cite{luo2019bag} introduced Bag of Tricks (BoT), which systematized an effective combination of existing techniques, such as batch normalization adjustment, learning rate warm-up, label smoothing, and a refined triplet loss.
With the rise of such high-performance ReID models, it has become possible to match people with high accuracy based on appearance similarity.

\subsection{Multi-Camera Tracking}
Multi-Camera Multi-Object Tracking (MCMOT) is a technique that uses multiple cameras to track objects over a wide area.
MCMOT can be broadly divided into two types depending on camera placement: overlapping and non-overlapping fields of view \cite{amosa2023multi}.

In non-overlapping camera views, ReID is particularly important.
He et al. \cite{he2019multi} combined visual features and spatio-temporal information to track vehicles across multiple cameras.
In addition, Bipin et al. \cite{gaikwad2021smart} proposed a method that combines person detection, tracking, and ReID, and focuses on real-time processing using edge devices.
On the other hand, in camera setups with overlapping views, many studies aim to improve tracking accuracy by combining ReID with position information \cite{yoshida2024overlap, xie2024robust}.
Yoshida et al. \cite{yoshida2024overlap} used global position data, pose-based image selection, and clustering-based re-identification to win the AI City Challenge 2024.
Xie et al. \cite{xie2024robust} combined geometric consistency with state-aware ReID correction, effectively reducing ID switches during occlusion and achieving high accuracy in real-time tracking.

However, many existing MCMOT methods have not fully addressed challenges specific to certain camera setups, such as image distortion caused by wide-angle views and inaccuracies in camera-to-camera alignment.
This study focuses on such tracking issues and presents practical solutions for real-world environments by comparing the use of position information and appearance features.

\section{ENVIRONMENT}
\subsection{Warehouse Environment}
This study is conducted in a logistics warehouse located in Aichi Prefecture, Japan.
As shown in Fig.\ref{fig:cam_config_1st_floor}, the warehouse has a large-scale camera system with more than 80 fixed cameras.
In this study, we used 19 of these ceiling-mounted H.View HV-800G2A5 cameras to capture the floor from a top-down view.
The video footage used in this study was recorded in Full HD at 5 fps.

\begin{figure}[t]
    \centering
    \includegraphics[width=0.45\textwidth]{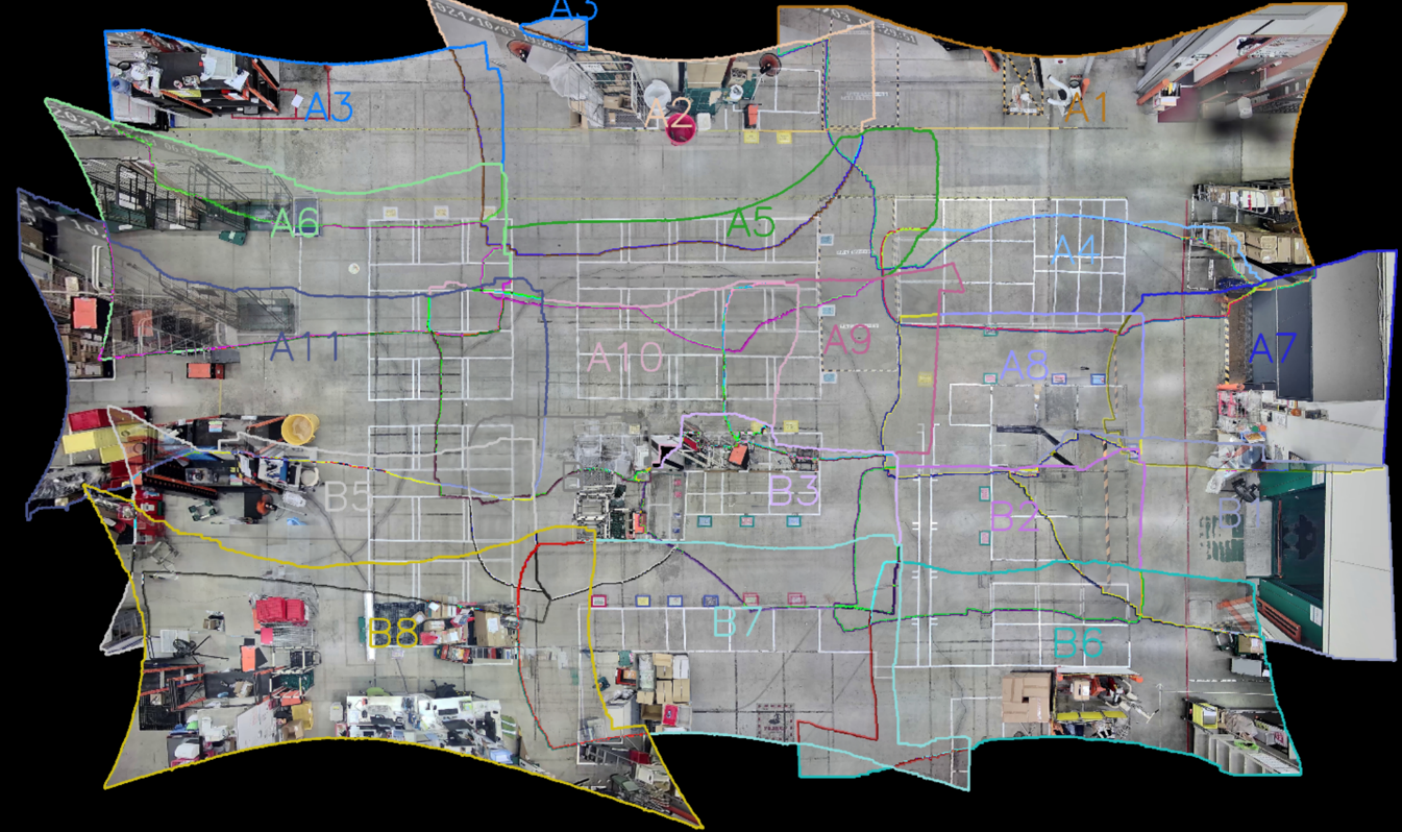}
    \caption{Camera alignment result}
    \label{fig:stitching_image}
\end{figure}

\subsection{Camera Placement and Multi-Camera Image Alignment}\label{sec:Stiching_matrics}
Due to installation constraints, we placed wide-angle cameras with a 110-degree field of view unevenly on the warehouse ceiling.
This required us to perform distortion correction and register the position of each camera.
We used the Double Sphere Model \cite{de0e1415c4d14bdcb102504f93c8602d}, which provides a good balance between speed and accuracy, for distortion correction.
However, after mounting the cameras on the ceiling, we observed some camera-specific distortion effects that remained unresolved, resulting in calibration challenges.
These unresolved issues are left for future work.

The distortion-corrected images were used for object detection and camera alignment.
To register camera positions, we used a color-mapped floor point cloud captured with the Leica's BLK2GO \cite{Leica_Geosystem}, and estimated the relative positions and orientations between cameras using keypoint matching with SuperPoint \cite{detone2018superpoint} and LightGlue \cite{lindenberger2023lightglue}.
Based on these correspondences, we derived projection matrices for converting detection results into global coordinates.
Fig.\ref{fig:stitching_image} shows the result of image alignment after distortion correction.
Although this alignment method achieves relatively high accuracy in the floor region, some positional misalignments still remain, and addressing them is a future challenge.

\section{METHOD}
\begin{figure*}[t]
    \centering
    \includegraphics[width=\textwidth]{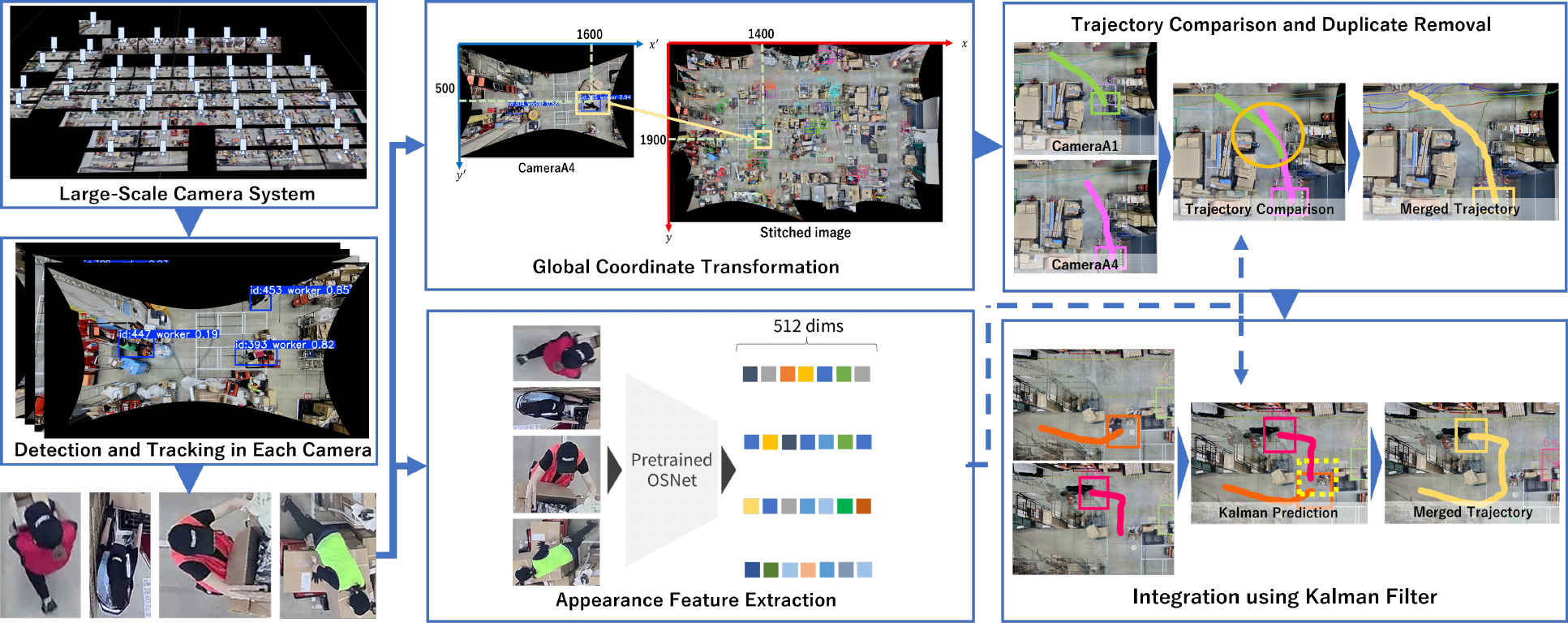}
    \caption{Framework of the proposed multi-camera tracking method}
    \label{fig:multicam_tracking_framework}
\end{figure*}

The framework of the proposed multi-camera tracking method is shown in Fig.\ref{fig:multicam_tracking_framework}.
It consists of five main processes: (1) worker detection and tracking in each camera, (2) appearance feature utilization, (3) global coordinate transformation of detection results, (4) trajectory comparison and duplicate removal, and (5) integration using a Kalman filter.
In the appearance feature utilization section, we detail both the extraction of appearance features and their application within Processes (4) and (5).

\begin{figure}[t]
    \centering
    \includegraphics[width=0.45\textwidth]{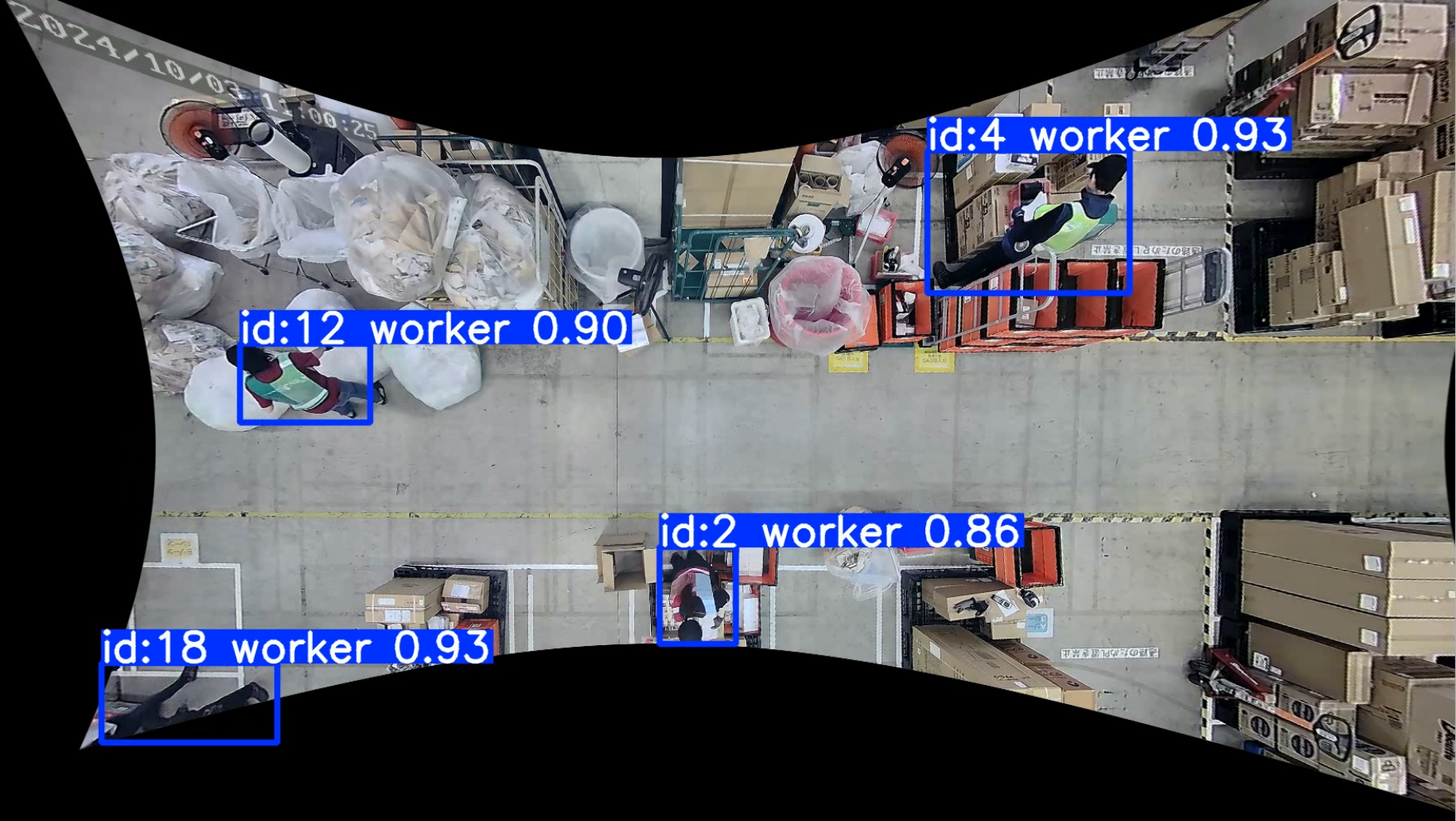}
    \caption{Example of tracking with a single camera}
    \label{fig:tracking_A2}
\end{figure}

\subsection{Worker Detection and Tracking in Each Camera}\label{sec:Detection_per_cam}
To detect and track workers in each camera, we use a combination of YOLOv8 (You Only Look Once) \cite{redmon2016you} and ByteTrack.
First, we apply fine‑tuned YOLOv8x model to detect workers in each frame.
Then, ByteTrack is used to associate detection results across frames and track each individual.
An example of single-camera tracking is shown in Fig.\ref{fig:tracking_A2}.
Each track represents the trajectory of a person and is assigned a unique ID.

\subsection{Appearance Feature Utilization (Comparison Method)}
\subsubsection{Appearance Feature Extraction}\label{sec:Extract_feature}
When tracking across cameras, using appearance features in addition to position information can improve tracking accuracy.
In this study, we use OSNet\_x1\_0 to extract appearance features of workers from each detected bbox in Section \ref{sec:Detection_per_cam}.
A fine-tuned model is used for feature extraction.

\begin{figure}[t]
    \centering
    \includegraphics[width=0.45\textwidth]{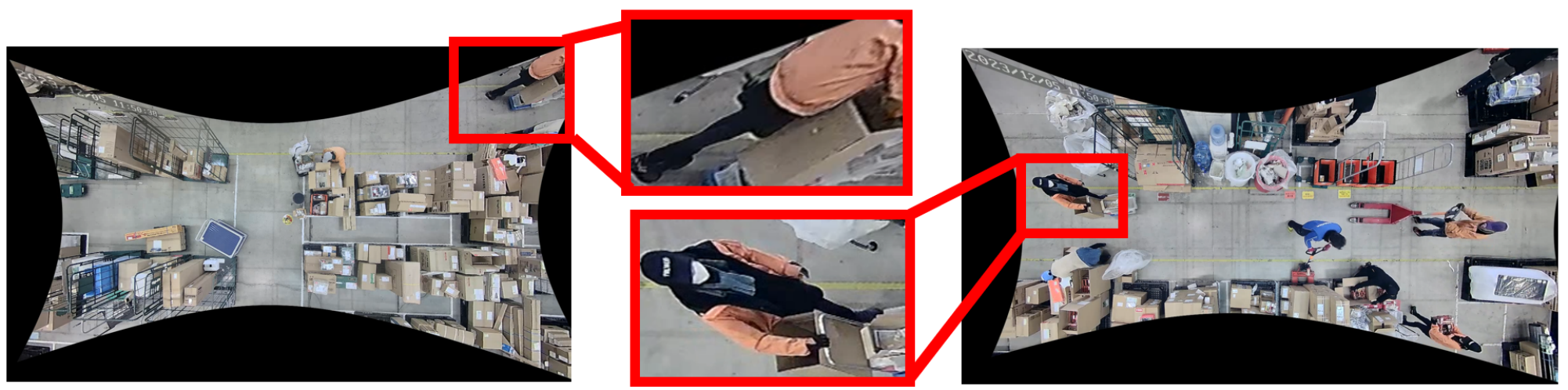}
    \caption{Variation in appearance of the same person}
    \label{fig:difference_feature_of_worker}
\end{figure}

\subsubsection{Comparison of Feature Usage Methods}\label{sec:Using_feature}
Appearance features can be useful for improving tracking accuracy.
However, since wide-angle cameras are used, the appearance of a person can vary significantly depending on their position in the image, as shown in Fig.\ref{fig:difference_feature_of_worker}.
In addition, the logistics warehouse is filled with a wide variety of objects, resulting in situations where only part of a worker's body is visible or where the worker is carrying tools or packages.
Such variations in appearance can lead to incorrect judgments when using appearance similarity for tracking.
Therefore, we compare two different methods of using appearance features and examine their effectiveness.

\paragraph{Simple Averaging}\label{sec:smoothing}
We compute the average of appearance features obtained from multiple detection results within the same track and use it as the representative feature of that track.
By averaging the features of detection results with the same tracking ID, we can reduce the effects of temporary appearance changes or cases where only some parts of the body are visible.

\begin{figure}[t]
    \centering
    \includegraphics[width=0.45\textwidth]{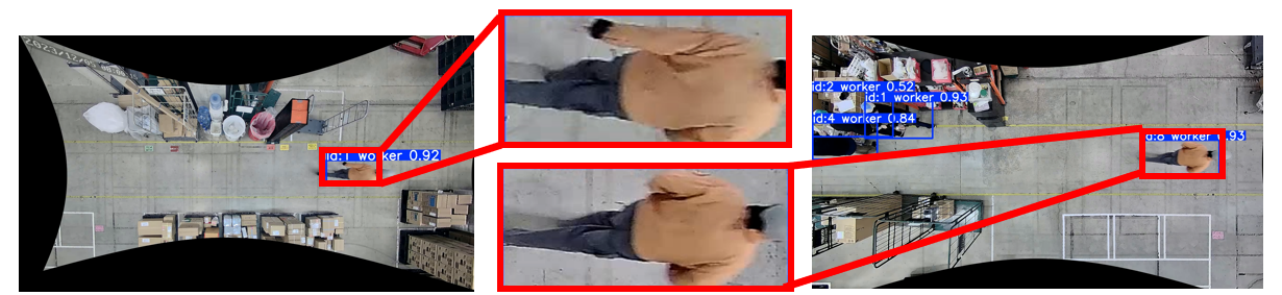}
    \caption{Example of detection results with similar positions and movement directions}
    \label{fig:similar_feature_of_worker}
\end{figure}

\paragraph{Position and Direction-Aware Method}\label{sec:v_p_feature}
As shown in Fig.~\ref{fig:similar_feature_of_worker}, detection results with similar positions and movement directions in the image tend to have similar appearance features, even across different cameras. These cases are also less affected by wide-angle distortion. 
To ensure a more reliable comparison of appearance features, we only compare detection results that are close in both position and movement direction.

Based on this idea, we compare two tracks A and B as follows. 
The detection results in track A are denoted as \(\{p_i\}_{i=1}^{N_A}\), and those in track B as \(\{q_j\}_{j=1}^{N_B}\). 
Each pair \((p_i, q_j)\) is referred to as a detection pair. The comparison procedure is as follows:

\begin{enumerate} 
  \item For each detection pair \((p_i, q_j)\), we calculate the distance $d_{ij} = \lVert c_{p_i} - c_{q_j} \rVert$ between the centers of their bboxes in the camera coordinate system. 
  If \(d_{ij}\) exceeds the distance threshold \(D_{\max}\), the detection pair is excluded. 
  Based on this distance, we assign a position weight $w_{pos}^{(i,j)}$, where closer pairs receive higher weights. 
  $\sigma_p$ is a parameter that controls the weighting scale.
        \[
          w_{pos}^{(i,j)} = \exp\!\Bigl(-\tfrac{d_{ij}^2}{\sigma_p^2}\Bigr)
        \]      
        
  \item Let the movement vector of \(p_i\) be \(\boldsymbol{v}_{p_i}\), and that of \(q_j\) be \(\boldsymbol{v}_{q_j}\). 
  Using the cosine similarity \(\cos_{\mathrm{vel}}^{(i,j)}\) between these vectors, we calculate the direction weight \(w_{\mathrm{vel}}^{(i,j)}\), which assigns higher values to pairs with more similar directions.  
        \[
          w_{vel}^{(i,j)} = \frac{1 + \cos_{vel}^{(i,j)}}{2}
        \]

  \item To give higher weights to detection pairs that are close in both position and movement direction, we compute the overall weight \(w^{(i,j)}\) as the product of the position weight \(w_{\mathrm{pos}}^{(i,j)}\) and the direction weight \(w_{\mathrm{vel}}^{(i,j)}\).  
        \[
          w^{(i,j)} = w_{pos}^{(i,j)} \times w_{vel}^{(i,j)}
        \]
  
  \item If the number of detection pairs is greater than or equal to a threshold \(M\), we select the top \(M\) pairs by weight \(w^{(i,j)}\), compute their appearance similarities, and take the average of the top 75\% as the final similarity.  
  This reduces the impact of outliers and emphasizes reliable comparisons.  
  On the other hand, if the number of detection pairs is less than \(M\), the purpose of comparing pairs that are close in position and movement direction may not be fulfilled, and the results may be more affected by outliers. 
  Therefore, in this case we use the value defined in Section \ref{sec:smoothing}.
\end{enumerate}

\begin{figure}[t]
    \centering
    \includegraphics[width=0.45\textwidth]{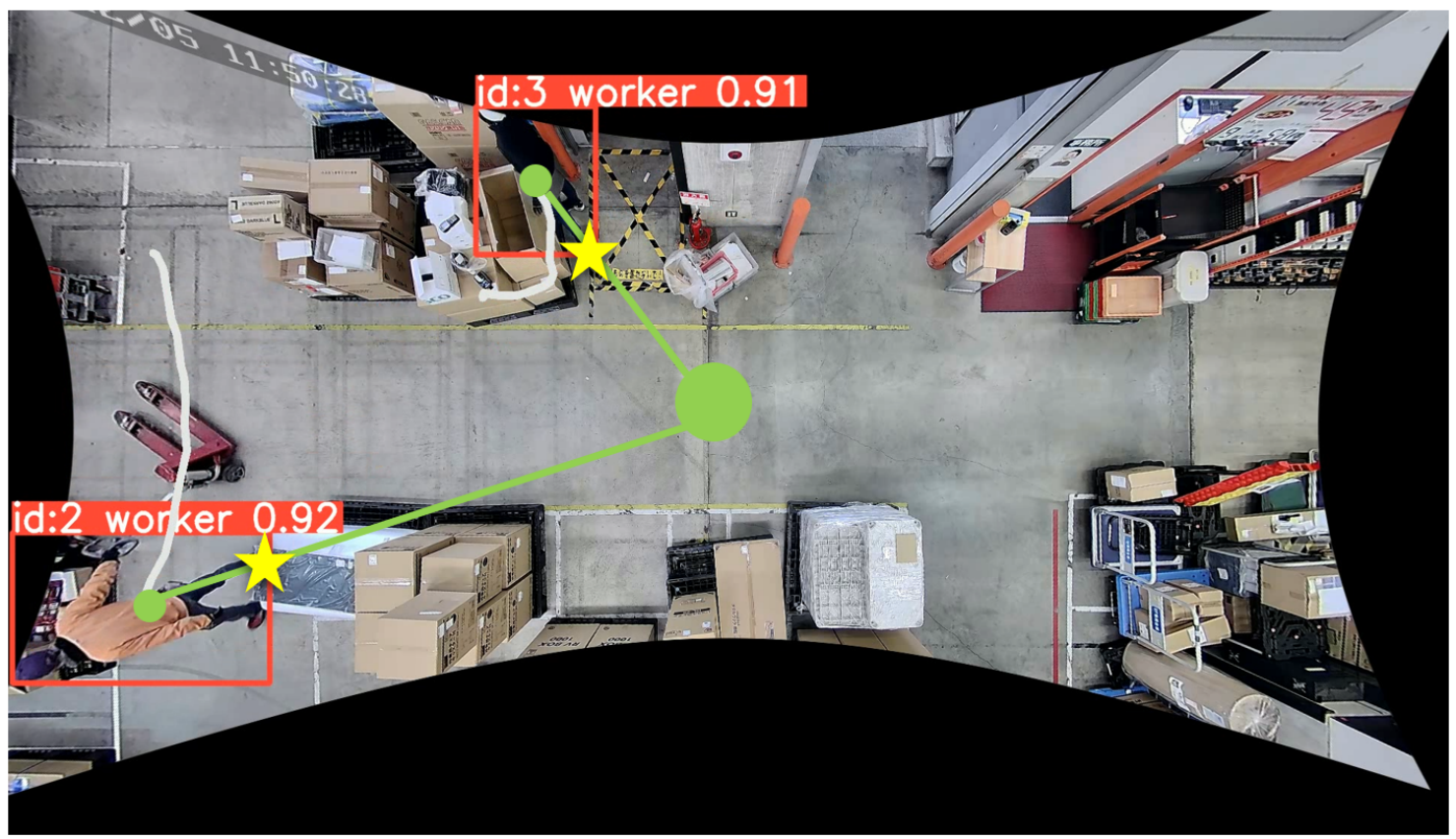}
    \caption{Estimation of foot position}
    \label{fig:calculate_leg_place}
\end{figure}

\subsection{Global Coordinate Transformation of Detection Results}\label{sec:leg_place}
To enable tracking across the entire receiving area of the warehouse, detection results from each camera are transformed into a global coordinate system.
For this purpose, we use the projection matrices constructed in Section \ref{sec:Stiching_matrics} to convert the local camera coordinates into global coordinates.
As described in Section \ref{sec:Stiching_matrics}, these projection matrices were obtained through floor surface feature matching.
Therefore, the floor is less affected by distortions and alignment errors.
In contrast, height-related information is more sensitive to such distortions and misalignments.
To address this, we compare two methods for converting detection results into global coordinates:
First, we use the center point of the detected bbox as the reference position. 
Second, as shown in Fig.\ref{fig:calculate_leg_place}, we estimate the foot position by finding the intersection between the line connecting the camera center and the center of the bbox, and one of the edges of the bbox.
Since the cameras in this study are mounted on the ceiling and look down toward the floor, a person's feet typically appear closer to the center of the image than their head.
Therefore, the intersection point between the line from the bbox center to the camera center and the bbox edges can be regarded as the point closest to the floor—i.e., the foot position.
By using this estimated foot position, the method reduces the effects of wide-angle distortion and misalignment between cameras, leading to more accurate tracking.

\begin{figure}[t]
    \centering
    \includegraphics[width=0.45\textwidth]{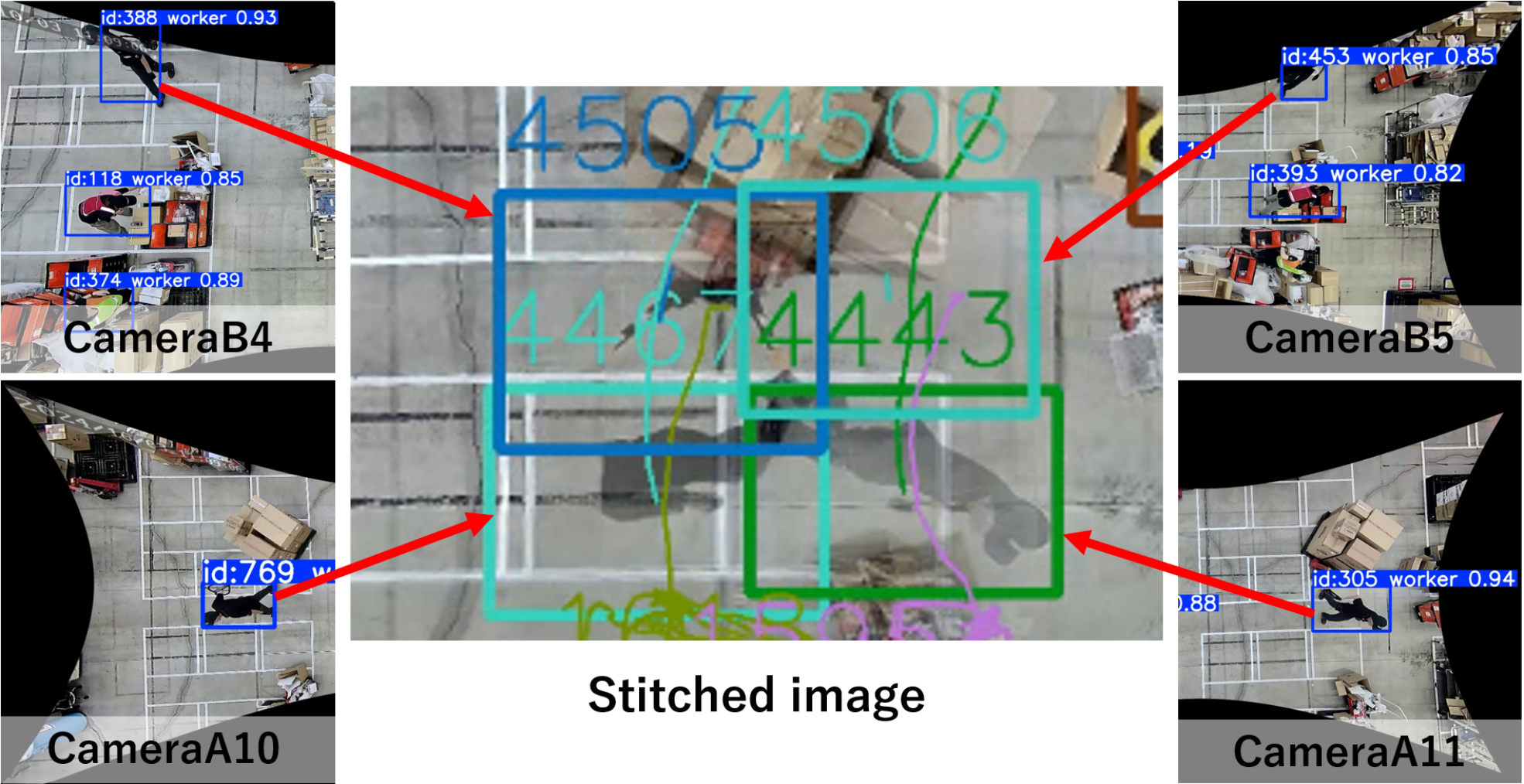}
    \caption{Example of duplicate detection}
    \label{fig:depulicate_image}
\end{figure}

\subsection{Trajectory Comparison and Duplicate Removal}\label{sec:process_duplicated}
As described in Section \ref{sec:Stiching_matrics}, due to overlapping fields of view, the same worker may be detected and tracked by multiple cameras, as shown in Fig.\ref{fig:depulicate_image}.
To address this, we compare the trajectories from each camera using the following elements and merge duplicated or fragmented tracks.

\begin{itemize}
  \item \textbf{Positional consistency:}  
  We evaluate the average Euclidean distance and the maximum distance between detection points of two tracks in overlapping frames.
  Tracks with close spatial proximity are considered as candidates for merging. 
  In addition, candidates for merging must come from different cameras.

  \item \textbf{Direction consistency:}  
  For the overlapping frame range, we compare the displacement vectors from start to end points of the two tracks using cosine similarity.  
  Only tracks with high similarity are considered for merging.

  \item \textbf{Appearance similarity:}  
  We use the appearance feature similarity described in Section \ref{sec:Using_feature} and tracks with high feature similarity are considered as merging candidates.
\end{itemize}

We assign the same tracking ID to tracks that satisfy all of the above conditions.  
After merging, if the same ID appears multiple times in a single frame, we keep only the detection result whose camera coordinate is closer to the center of the image, in order to reduce the effect of distortion.

\subsection{Integration using Kalman Filter}\label{sec:kalmanfilter}
In Section \ref{sec:process_duplicated}, we merged tracks based on the Euclidean distance and appearance similarity in overlapping frames.  
However, when there are no overlapping frames between two tracks, comparison becomes difficult, and the tracking results may become fragmented.
To address this, we apply the Kalman filter to each track and use the predicted position, movement direction, and appearance features to enable reliable merging even when there are frame gaps between detection results.
Specifically, we predict the next position using the Kalman filter from the end of a track, and compute the cosine similarity between the predicted movement direction and the initial movement direction of another track.  
We also calculate the appearance similarity between the tracks using the method described in Section \ref{sec:Using_feature}.  
If the appearance similarity is high, we allow a larger distance and a wider angle difference when evaluating spatial and directional consistency.  
If the appearance similarity is low, we apply stricter conditions for merging.  
Finally, if the predicted position from the Kalman filter is close enough to the starting point of the next track, and the cosine similarity of their movement directions is above a threshold, the tracks are merged as the same person.


\section{EXPERIMENT}
\subsection{Data Used in the Experiment}\label{sec:Environment_evaluation}
In this experiment, we used 30-minute videos recorded at 5 fps and Full HD from 19 cameras.  
The camera placement follows the configuration described in Section \ref{sec:Stiching_matrics}.

\subsection{Experimental Setup}
\subsubsection{Training the Object Detection Model}\label{sec:train_yolo}
In this experiment, we used the fine‑tuned YOLOv8x model for object detection.  
For training, we prepared a dataset using video footage recorded at a different time from the one described in Section \ref{sec:Environment_evaluation}.  
The dataset was created using efficient annotation methods \cite{kano2024composite, higashiura2024semi, mori2025efficient}, as well as manual annotation.  
In the manual annotation process, each target object was manually segmented by drawing bboxes and labeled with appropriate class names.

Table \ref{tab:detector_dataset_stats} shows the number of data used for training and validation.  
The input size of the model was set to 640×640.

\begin{table}[t]
  \centering
  \caption{Dataset used for training the object detection model}
  \label{tab:detector_dataset_stats}
  \begin{tabular}{ccc} \hline
    Dataset & Number of Images & Number of Objects \\ \hline
    Train      & 11,459  & 29,473 \\
    Validation & 1,373   & 3,629  \\ \hline
  \end{tabular}
\end{table}

\subsubsection{Training the Appearance Feature Extraction Model}\label{sec:train_osnet}
To extract appearance features of workers, we used the OSNet\_x1\_0 model and fine-tuned it using a custom-built dataset.  
This dataset was created from 10-minute videos recorded at a different time from the one described in Section 
 \ref{sec:Environment_evaluation}.

First, workers were detected in each frame using YOLO, and person images were cropped using the detected bboxes.  
Then, tracking was performed using the proposed method.  
Since the tracking results contained some errors, we manually corrected all tracking IDs to create accurate annotations that correctly link the same person across multiple cameras.
To avoid including too many similar samples from nearby frames, we extracted training data every 20 frames.
The final dataset consisted of 3,029 cropped images of 40 individuals captured by 19 cameras.  
All images were resized to 256×128 pixels.  
The model was trained with a batch size of 64 for up to 250 epochs.

\subsubsection{Evaluation Metrics and Ground Truth}
We used the following three metrics to evaluate the tracking results:

\begin{itemize}
  \item \textbf{Higher Order Tracking Accuracy (HOTA)}\cite{luiten2020IJCV}:  
  A metric that balances detection accuracy and association accuracy, providing a comprehensive evaluation of overall tracking performance.  
  \item \textbf{ID F1-score (IDF1)}\cite{ristani2016performance}:  
  A metric that evaluates how well the same ID is maintained for each object.  
  It focuses specifically on ID consistency.  
  \item \textbf{Multiple Object Tracking Accuracy (MOTA)}\cite{bernardin2008evaluating}:  
  A metric that considers false positives, false negatives, and ID switches.  
  It is sensitive to tracking errors and provides an overall measure of tracking accuracy.
\end{itemize}

These metrics are widely used in tracking evaluation and are all originally evaluated on a scale from 0 to 1, where values closer to 1 indicate better performance.  
In this paper, we multiply the metric values by 100, and present them on a 0 to 100 scale.  
We used TrackEval \cite{luiten2020trackeval} to compute the evaluation metrics.

The ground truth data was created by manually correcting and supplementing the tracking results produced by our proposed method.  
We carefully fixed ID switches, missed detections, and false positives to ensure high-quality annotations.

\subsection{Evaluation Experiments}
\subsubsection{Comparison Settings}
To evaluate the contribution of each factor to improving tracking accuracy, we conducted a comparative analysis using different combinations of coordinate types (foot position or detection center) and appearance feature usage methods (none, simple averaging, or position and direction-aware).



\subsubsection{Experimental Procedure}
To evaluate how each component of the proposed method affects tracking accuracy, we conducted comparative experiments based on our framework.
First, we performed object detection and tracking on the camera footage described in Section \ref{sec:Environment_evaluation}.  
For detection, we used the YOLOv8x model fine-tuned as described in Section \ref{sec:train_yolo}.  
Next, appearance features were extracted using the OSNet\_x1\_0 model described in Section \ref{sec:train_osnet}.  
The detection results were then transformed into the global coordinate system using the projection matrices described in Section \ref{sec:leg_place}, based on either the center of the bbox or the estimated foot position.
Following this, duplicate and fragmented tracks were processed as described in Section \ref{sec:process_duplicated}.  
Tracks were merged if the average distance between their global coordinates was within 130 pixels and all of the following conditions were met.
The threshold values used in these processes were selected by evaluating multiple candidate values and adopting the configuration that achieved the highest tracking scores.

\begin{itemize}
  \item In all overlapping frames, the distance between the two objects is less than 300 pixels.
  \item The cosine similarity of movement directions is greater than 0.8.
  \item The cosine similarity of appearance features (Section \ref{sec:Using_feature}) is greater than 0.85.
\end{itemize}

Finally, the merging process described in Section \ref{sec:kalmanfilter} was applied.  
Two tracks were merged if all the following conditions were satisfied:

\begin{itemize}
  \item The frame gap between the two tracks is less than 10 frames.
  \item The cosine similarity between the predicted movement direction vector of one track and the initial movement direction vector of the other is greater than 0.8.
  \item The predicted position is within a threshold distance (130 pixels) from the start of the next track.  
  If the appearance similarity is greater than 0.85, the threshold is doubled.  
  If the similarity is low (less than 0.5), the threshold is reduced to half.
\end{itemize}

Other parameters (Section \ref{sec:Using_feature}) were set as $\sigma_p = 500$, $D_{\max} = 540\ \mathrm{px}$ and $M = 8$.


\begin{table*}[t]
  \centering
  \caption{Tracking evaluation results under different experimental conditions}
  \label{tab:experiment_conditions}
  \begin{tabular}{lccc} \hline
    \textbf{Condition} & \textbf{HOTA} & \textbf{IDF1} & \textbf{MOTA} \\ \hline
    (1) Bbox center only & 39.6 & 39.7 & 65.7 \\
    (2) Bbox center + Appearance features (simple averaging) & 48.5 & 49.7 & 77.0 \\
    (3) Bbox center + Appearance features (position and direction-aware) & 46.9 & 47.0 & 75.4 \\ 
    (4) Foot position only & 49.1 & 50.2 & 78.9 \\
    (5) Foot position + Appearance features (simple averaging) & \textcolor{red}{\textbf{51.0}} & \textcolor{red}{\textbf{54.7}} & \textcolor{red}{\textbf{79.7}} \\
    (6) Foot position + Appearance features (position and direction-aware) & \textcolor{blue}{50.8} & \textcolor{blue}{54.5} & \textcolor{blue}{79.2} \\ \hline
  \end{tabular}
\end{table*}

\begin{figure*}[t]
    \centering
    \includegraphics[width=0.90\textwidth]{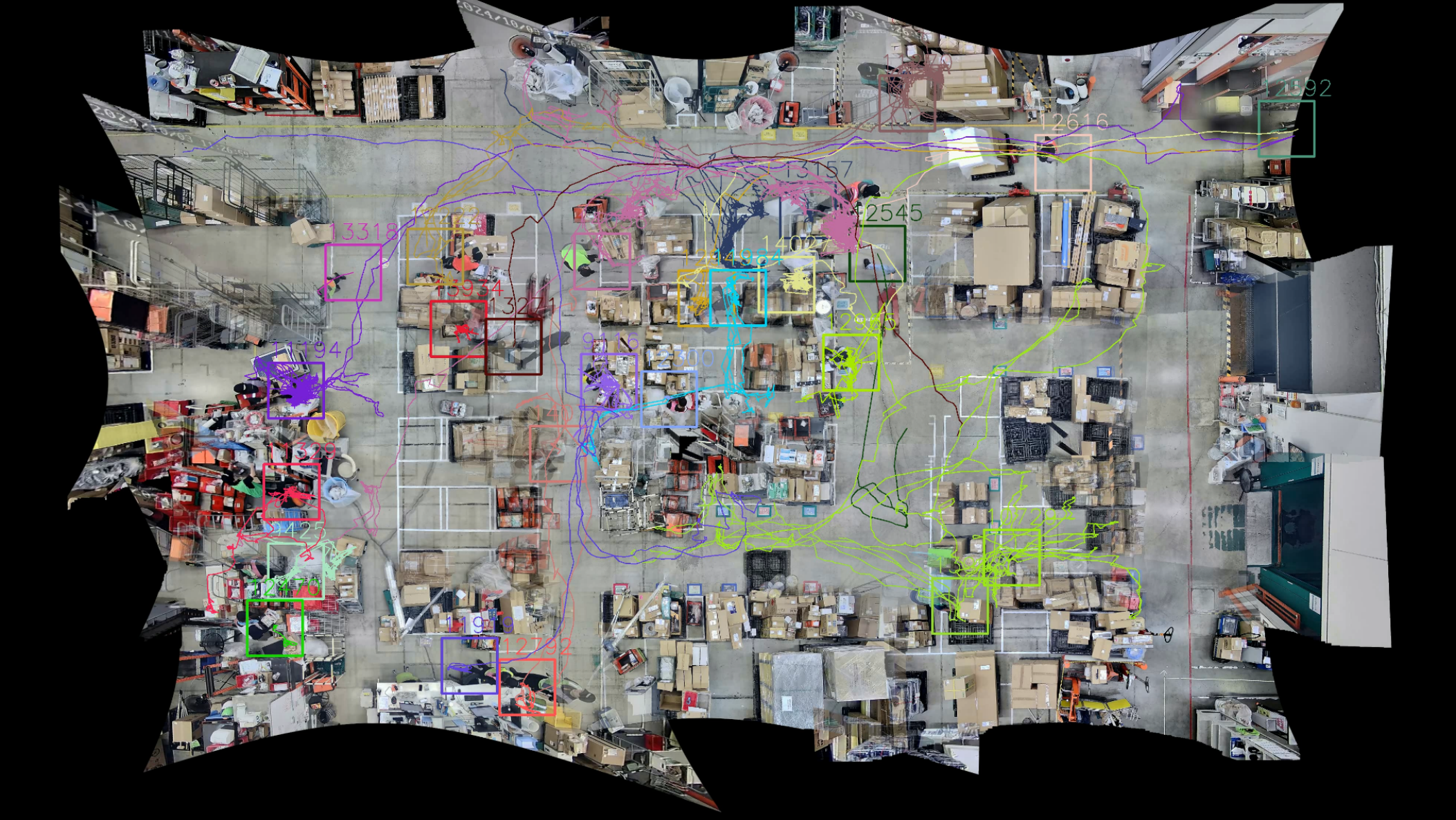}
    \caption{Example of tracking results}
    \label{fig:multicam_result}
\end{figure*}

\subsection{Results and Discussion}
Table \ref{tab:experiment_conditions} presents the tracking results under six different conditions, and Fig.\ref{fig:multicam_result} shows the tracking outcome under the highest-performing setting.

We first discuss the effect of coordinate type.  
Without using appearance features, the HOTA, IDF1, and MOTA scores for the bbox center were 39.6, 39.7, and 65.7, respectively, while those for the foot position were significantly higher at 49.1, 50.2, and 78.9.  
This suggests that using foot position helps reduce the impact of distortion and alignment errors caused by wide-angle cameras.

Next, we discuss the effect of appearance feature usage.  
In both coordinate settings, the use of appearance features improved identification and matching accuracy.  
With simple averaging, the HOTA, IDF1, and MOTA scores were 51.0, 54.7, and 79.7 for the foot position, and 48.5, 49.7, and 77.0 for the bbox center.  
When position and direction were considered, the scores were 50.8, 54.5, and 79.2 for the foot position, and 46.9, 47.0, and 75.4 for the bbox center.  
In all cases, performance—especially IDF1—improved with the use of appearance features, and the highest accuracy was achieved with the combination of foot position and simple averaging.  
Interestingly, the simple averaging method slightly outperformed the position and direction-aware method.
This suggests that it is important to reduce the effects of short-term changes in appearance and situations where only part of the body is visible.

Based on these results, it is clear that using foot position greatly improves the localization and association accuracy of tracking.
In addition, the use of appearance features helps make identity matching more stable and improves overall tracking performance.
In particular, the combination of foot position and simple averaging of appearance features achieved the highest accuracy.
This suggests that for accurate tracking in multi-view environments, it is important to properly combine spatial and appearance information.


\section{CONCLUSION and FUTURE WORK}

In this study, we proposed a method to improve multi-camera worker tracking performance, particularly in environments equipped with wide-angle cameras.
We particularly compared two types of location representations: the center of the detected bbox and the estimated foot position.
This comparison showed that using foot positions helps reduce the effects of image distortion and camera misalignment.
Under conditions where appearance features were not used, using foot positions led to relative improvements of 24\% in HOTA, 26\% in IDF1, and 20\% in MOTA.
In addition, we evaluated two different methods for using appearance features extracted by OSNet: a simple averaging approach and a method that considers both position and movement direction.
This analysis clarified the strengths and limitations of each method and how they affect tracking performance.
To demonstrate the effectiveness of the proposed method, we applied it to video data from 19 wide-angle cameras installed in the logistics warehouse, and confirmed that the appropriate integration of spatial and appearance information contributes to improved tracking accuracy.

One of the main reasons for tracking failures is misalignment between cameras.
In this study, we used foot positions to reduce the impact of misalignment and achieve more stable coordinate transformation.
However, using foot positions alone does not completely solve the problem.
Misalignment can still affect positional consistency when merging tracking results across cameras, sometimes resulting in assigning different IDs to the same person.
In some cases, the lower body of a worker is occluded by objects, and only the upper body is detected. As a result, the foot position cannot be obtained.

To address these issues, we plan to improve the merging process by incorporating temporal information and motion consistency, which may help reduce the effects of misalignment.
We also consider introducing a complementary method to estimate foot positions from other body parts, such as the upper body, when the feet are not visible.
Furthermore, we aim to improve tracking robustness by integrating external data sources such as beacon data \cite{kano2022smartphone}.

By pursuing these directions, we aim to further improve the accuracy and reliability of worker tracking in the logistics warehouse, and contribute to more efficient operations.

\section*{ACKNOWLEDGMENT}
This work was partially supported by NEDO (JPNP23003), JSPS KAKENHI (JP22K18422), and TRUSCO Nakayama Corporation.
\bibliographystyle{IEEEtran}
\bibliography{ref}

\end{document}